\documentclass{ecai}

\usepackage{latexsym}
\usepackage{amssymb}
\usepackage{amsmath}
\usepackage{amsthm}
\usepackage{booktabs}
\usepackage{enumitem}
\usepackage{graphicx}
\usepackage{color}

\usepackage{enumitem}
\usepackage{float}
\usepackage{comment}
\usepackage{svg}

\newcommand{\BibTeX}{B\kern-.05em{\sc i\kern-.025em b}\kern-.08em\TeX}
\newcommand{\abs}[1]{\left| #1 \right|}
\newcommand{\model}{Diff-FSA}

\begin{document}

\begin{frontmatter}

\paperid{2307}

\title{GraphFSA: A Finite State Automaton Framework for Algorithmic Learning on Graphs}

\author[A]{\fnms{Florian}~\snm{Grötschla}\footnote{Equal contribution.}}
\author[A]{\fnms{Joël}~\snm{Mathys}\footnotemark}
\author[A]{\fnms{Christoffer}~\snm{Raun}} 
\author[A]{\fnms{Roger}~\snm{Wattenhofer}} 

\address[A]{ETH Zurich}

\begin{abstract}
Many graph algorithms can be viewed as sets of rules that are iteratively applied, with the number of iterations dependent on the size and complexity of the input graph. Existing machine learning architectures often struggle to represent these algorithmic decisions as discrete state transitions. Therefore, we propose a novel framework: GraphFSA (Graph Finite State Automaton). GraphFSA is designed to learn a finite state automaton that runs on each node of a given graph. We test GraphFSA on cellular automata problems, showcasing its abilities in a straightforward algorithmic setting. For a comprehensive empirical evaluation of our framework, we create a diverse range of synthetic problems. As our main application, we then focus on learning more elaborate graph algorithms. Our findings suggest that GraphFSA exhibits strong generalization and extrapolation abilities, presenting an alternative approach to represent these algorithms. 

\end{abstract}

\end{frontmatter}

\section{Introduction}
While machine learning has made tremendous progress, machines still have trouble generalizing concepts and extrapolating to unseen inputs. Large language models can write spectacular poems about traffic lights, but they still fail at multiplying two large numbers. They do not quite understand the multiplication algorithm since they do not have a good representation of algorithms. We want to teach machines some level of ``algorithmic thinking.'' Given some unknown process, can the machine distill what is going on and then apply the same algorithm in another situation? This paper concentrates on one of the simplest processes: finite state automata (FSA). An FSA is a basic automaton that jumps from one state to another according to a recipe. FSAs are the simplest, interesting version of an algorithm. However, if we assemble many FSAs in a network, the result is remarkably powerful regarding computation. Indeed, the simple Game of Life is already Turing-complete. \\

Building on the work of \citet{grattarola2021learning} and \citet{Marr_2009}, this paper presents GraphFSA (Graph Finite State Automaton), a novel framework designed to learn a finite state automata on graphs. GraphFSA extracts interpretable solutions and provides a framework to extrapolate to bigger graphs, effectively addressing the inherent challenges associated with conventional methods. To better understand the capabilities of GraphFSA, we evaluate the framework on various cellular automata problems. In this controllable setting, we can test the model's abilities to extrapolate and verify that GraphFSA learns the correct rules. In addition, we introduce a novel dataset generator to create problem instances with known ground truth for the evaluation of GraphFSA models. 
Subsequently, we consider more elaborate graph algorithms to evaluate GraphFSA performance on more complex graph problems. Our paper makes the following key contributions:

\begin{itemize}[leftmargin=*, topsep=0pt]
\itemsep0em 
\item We introduce GraphFSA, an execution framework designed for algorithmic learning on graphs, addressing the limitations of existing models in representing algorithmic decisions as discrete state transitions and extracting explainable solutions.
\item We present GRAB: The \textbf{GR}aph \textbf{A}utomaton \textbf{B}enchmark, a versatile dataset generator to facilitate the systematic benchmarking of GraphFSA models across different graph sizes and distributions. \footnote{The code is available at https://github.com/ETH-DISCO/graph-fsa.}
\item We present \model, one specific approach to train models for the GraphFSA framework and use GRAB to compare its performance to a variety of baselines, showcasing that our model offers strong generalization capabilities.
\item We provide a discussion and visualization of GraphFSA's functionalities and model variations. We provide insights into its explainability, limitations, and potential avenues for future work, positioning it as an alternative in learning graph algorithms.
\end{itemize}

\section{Related work}
\paragraph{Finite state automaton.}
A Finite State Automaton (FSA), also known as a Finite State Machine (FSM), is a computational model used to describe the behavior of systems that operates on a finite number of states. It consists of a set of states, a set of transitions between these states triggered by input symbols from a finite alphabet, and an initial state. FSAs are widely employed in various fields, including computer science, linguistics, and electrical engineering, for tasks such as pattern recognition~\citep{Lucchesi1993ApplicationsOF}, parsing, and protocol specification due to their simplicity and versatility in modeling sequential processes.
Additionaly, there is a growing interest in combining FSA with neural networks to enhance performance and generalization, a notion supported by studies by ~\citet{Pigem2013LearningFM} on learning FSAs. Additional research by \citet{mordvintsev2022diff_fsm} shows how a differentiable FSA can be learned. 

\begin{figure*}[t]
\begin{center}
      \centering
    \includegraphics[scale=0.9]{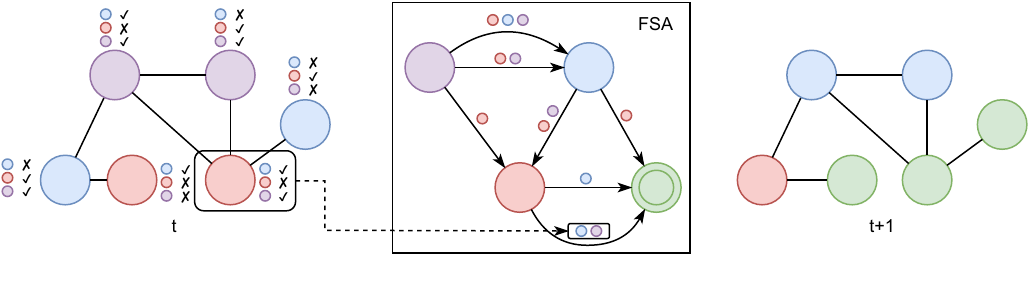}
\caption{Illustration of the GraphFSA framework: Each node has its own state, represented by its color. Furthermore, each node runs the same Finite State Automaton, determining its next state depending on the neighborhood information. On the left, an example graph is shown with the aggregation next to it. In this example, the aggregator can distinguish if there is a node of a certain color/state in the neighborhood. In general, the aggregator can be any function that maps a state multiset to a finite set of values.
State transitions are determined by the FSA depicted in the middle, which chooses chooses the next state based on the states appearing in the neighborhood and the old state. 
An example of the transition that is taken by the red node on the bottom right of the graph is shown. 
Note that the green state is final and does not have any outgoing transitions.}   
\label{fig:graphfsa-visualization}
\end{center}    
\end{figure*}

\paragraph{Cellular automaton.}
Cellular Automata (CA) are discrete computational models introduced by John von Neumann in 1967~\citet{Neumann1967TheoryOS}. They consist of a grid of cells, where each cell is in a discrete state and evolves following transition rules based on the states of neighboring cells. In our context, learning these transition functions from data and modeling the appropriate domain is of particular interest. Notably, recent work by \citet{10.5555/2987061.2987139} trains a Neural Network (NN) to learn 1D and 2D CAs, while further advancements are made with Neural Cellular Automata (NCA) introduced by \citet{48963}, and the Graph Cellular Automata (GCA) proposed by \citet{Marr_2009}, operating on graph structures with diverse neighborhood transition rules. Additionally, \citet{grattarola2021learning} introduce Graph Neural Cellular Automata (GNCA), focusing on learning a GCA with continuous state spaces.
GNCA mainly differs from our work in that it uses continuous instead of discrete state spaces and only learns a single transition step at a time, while we can extract rules from observations over multiple steps.
Lastly, the study by~\citet{johnson2020learning} investigates relationships defined by paths on a graph accepted by a finite-state automaton. 
Notably, this execution mechanism differs significantly from our methodology, serving the distinct purpose of adding edges on accepted paths. Unlike the synchronous execution of multiple finite-state automata in GraphFSA, here, the single automaton functions more like an agent traversing the graph. Consequently, the tasks the authors tackle, such as identifying additional edges in grid-world environments and abstract syntax trees (ASTs), differ from ours. It is crucial to highlight that this model only augments edges and does not predict node values.

\paragraph{Graph neural networks.}

Graph Neural Networks (GNNs) were introduced by \citet{4700287} and have gained significant attention. Various GNN architectures, such as Graph Convolutional Networks \citep{kipf2017semisupervised} and Graph Attention Networks \citep{veličković2018graph}, have been developed. A crucial area of GNN research is understanding their theoretical capabilities, which is linked to the WL color refinement algorithm that limits the conventional GNNs' expressiveness \citep{weisfeiler1968reduction,xu2019powerful,papp2022theoretical}. The WL algorithm proceeds in rounds where node colors are relabeled based on their own color and the colors of nodes in the neighborhood. To achieve optimal expressiveness, message-passing GNNs must run the layers an adequate number of rounds to match the computational prowess of WL. The necessity of augmenting the number of rounds has been emphasized for algorithmic graph problems \citep{loukas2020graph}. However, failing to adhere to the minimum rounds required may cause a GNN to fall short of solving or approximating specific problems, as shown by \citet{pmlr-v101-sato19a}. 

Furthermore, enhancing the interpretability and visualization of GNNs has become an essential part of GNN research. The development of GNNExplainer \citep{ying2019gnnexplainer} marked a significant stride in generating explanations for Graph Neural Networks. Similarly, \citet{8371286} delved into visual analytics in deep learning, providing an interrogative survey for future developments in this domain. Another approach towards explainability is understanding specific examples proposed by \citet{9811416}. Moreover, the GraphChef architecture introduced by \citet{muller2023graphchef}, offers a comprehensive understanding of GNN decisions by replacing layers in the message-passing framework with decision trees. GraphChef uses the gumbelized softmax from \citet{jang2017categorical} to learn a discrete state space, enabling the extraction of decision trees. 

\paragraph{Algorithmic learning.}
Algorithm learning aims to learn an underlying algorithmic principle that allows generalization to unseen and larger instances. As such, one of the primary goals is to achieve extrapolation. Various models have been investigated for extrapolation, including RNNs capable of generalizing to inputs of varying lengths \citep{963769}. Specifically, in our work, we want to learn graph algorithms. For extrapolation, we want our learned models to scale to graphs that are more complex and larger than those used during training. Graph Neural Networks (GNNs) have been extensively applied to solve these algorithmic challenges~\citep{selsam2019learning,Veli_kovi__2021,travellingGNN}. Recent studies by \citet{tang2020scaleinvariant}, \citet{veličković2022clrs}, and \citet{ibarz2022generalist} have emphasized improving extrapolation to larger graphs in algorithmic reasoning problems. However, most of the work focuses only on slightly larger graphs than those in the training set. Scaling to much larger graphs, as studied by \citet{AlgorithmicRecurrentGNN2022}  remains challenging. 

\section{The GraphFSA framework}
\label{sec:graphfsa}

We present GraphFSA, a computational framework designed for executing Finite State Automata (FSA) on graph-structured data. Drawing inspiration from Graph Cellular Automata and Graph Neural Networks, GraphFSA defines an FSA at each node within the graph. This FSA remains the same across all nodes and encompasses a predetermined set of states and transition values. While all nodes abide by the rules of the same automaton, the nodes are usually in different states. As is customary for FSAs, a transition value and the current state jointly determine the subsequent state transition. In our approach, transition values are computed through the aggregation of neighboring node states. As the FSA can only handle a finite number of possible aggregations, we impose according limitations on the aggregation function.
For the execution of the framework, nodes are first initialized with a state that represents their input features. The FSA is then executed synchronously on all nodes for a fixed number of steps.
Each step involves determining transition values and executing transitions for each node to change their state. Upon completion, each node reaches an output state, facilitating node classification into a predetermined set of categorical values. A visual representation of the GraphFSA model is depicted in Figure \ref{fig:graphfsa-visualization}.

\subsection{Formal definition}
\label{sec:formal-def-graphfsa}
More formally, the GraphFSA $\mathcal{F}$ consists of a tuple $(\mathcal{M}, \mathcal{Z}, \mathcal{A}, \mathcal{T})$. $\mathcal{F}$ is applied to a graph $G = (V, E)$ and consists of a set of states $\mathcal{M}$, an aggregation $\mathcal{A}$ and a transition function $\mathcal{T}$. At time $t$, each node $v\in V$ is in state $s_{v, t} \in \mathcal{M}$. In its most general form, the aggregation $\mathcal{A}$ maps the multiset of neighboring states to an aggregated value $a \in \mathcal{Z}$ of a finite domain. 

\begin{align}
    a_{v,t} = \mathcal{A}(\{\!\{ s_{u, t} \mid u \in N(v)\}\!\})
\end{align}

Here $\{\!\{\}\!\}$ denotes the multiset and $N(v)$ the neighbors of $v$ in $G$. At each timestep $t$, the transition function $\mathcal{T}: \mathcal{M} \times \mathcal{Z} \to \mathcal{M}$ takes the state of a node $s_{v,t}$ and its corresponding aggregated value $a_{v,t}$ and computes the state for the next timestep. 

\begin{align}
    s_{v, t+1} = \mathcal{T}(s_{v, t}, a_{v, t})
\end{align}

For notational convenience, we also define the transition matrix $T$ of size $\abs{\mathcal{M}}\times \abs{\mathcal{Z}}$ where $T_{m, a}$ stores $\mathcal{T}(m,a)$. Moreover, we introduce the notion of a state vector $s_v^t$ for each node $v \in V$ at time $t$, which is a one-hot encoding of size $\abs{\mathcal{M}}$ of the node's current state.

\paragraph{Aggregation functions.}
The transition value $a$ for node $v$ at time $t$ is directly determined by aggregating the multi-set of states from all neighboring nodes at time $t$. 
The aggregation $\mathcal{A}$ specifies how the aggregated value is computed from this neighborhood information. 
Note that this formulation of the aggregation $\mathcal{A}$ allows for a general framework in which many different design choices can be made for a concrete class of GraphFSAs. Throughout this work, we will focus on the \textit{counting} aggregation and briefly touch upon the \textit{positional} aggregation. 

\textit{Counting aggregation:}
The aggregation aims to count the occurrence of each state in the immediate neighborhood. However, we want the domain $\mathcal{Z}$ to remain finite. Note that due to the general graph topology of $G$, the naive count could lead to $\mathcal{Z}$ growing with $n$ the number of nodes. 
Instead, we take inspiration from the distributed computing literature, specifically the Stone Age Computing Model by \citet{emek2012stone}. Here, the aggregation is performed according to the one-two-many principle, where each neighbor can only distinguish if a certain state appears \texttt{once}, \texttt{twice}, or \texttt{more than twice} in the immediate neighborhood. Formally, we can generalize this principle using a bounding parameter $b$, which defines up to what threshold we can exactly count the neighbor states. The simplest mode would use $b = 1$, i.e., where we can distinguish if a state is part of the neighborhood or not. Note that this is equivalent to forcing the aggregation to use a set instead of a multi-set. For the general bounding parameter $b$, we introduce the threshold function $\sigma$, which counts the occurrences of a state in a multi-set.
\begin{align}
    \sigma(m, S) = \min(b, \abs{ \{\!\{s \mid s = m, s \in S\}\!\} }) 
\end{align}
Therefore, the aggregated value $a_{v,t}$ will be an $\abs{\mathcal{M}}$ dimensional vector of the following form, where the $m$-th dimension corresponds to the number of occurrences of state $m$. 
\begin{align}
    (a_{v,t})_m = \sigma(m, \{\!\{s_{u,t} \mid u \in N(v)\}\!\}) 
\end{align}

Note that there is a tradeoff between expressiveness and model size when choosing a higher bounding parameter $b$. 
For a model with $|\mathcal{M}|$ number of states there are ${(b+1)}^{|\mathcal{M}|}$ many transition values for each state $m \in \mathcal{M}$.

\textit{Positional aggregation:} This aggregation function takes into account the spatial relationship between nodes in the graph. If the underlying graph has a direction associated with each neighbor, i.e., in a grid, we assign each neighbor to a distinct slot within a transition value vector $a_{v,t}$. This generalizes to graphs of maximum degree $d$, where each of the d entries in $a_{v,t}$ corresponds to a state of a neighboring node.

\textit{Average threshold-based aggregation} An alternative aggregation operator utilizes the average value, which allows us to detect whether a specific threshold $t$ of neighbors is in a particular state. We compute the value to threshold by dividing the state sum by the neighborhood size of a node v. That, for example, allows us to detect whether the majority (e.g., 0.5) of neighbors are in a specific state. 

\paragraph{Starting and final states.}
We use starting states $S \subseteq \mathcal{M}$ of the FSA to encode the discrete set of inputs to the graph problem. These could be the starting states of a cellular automaton or the input to an algorithm (e.g., for BFS, the source node starts in a different state from the other nodes). 
Final states $F \subseteq \mathcal{M}$ are used to represent the output of a problem. In node prediction tasks, we choose one final state per class. Opposed to other states, it is not possible to transition away from a final state, meaning that once such a state is reached, it will never change.

\begin{figure}[t]
\begin{center}
      \centering
    \includegraphics[width=\linewidth]{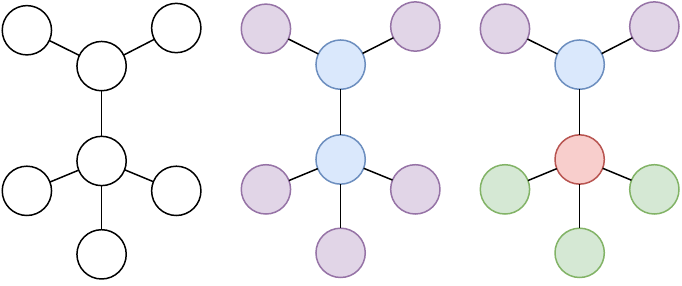}
\caption{Original topology (left), nodes that GraphFSA can distinguish with 2+ aggregation (i.e., distinguish neighbors as \texttt{0}, \texttt{1} or \texttt{more than 2}) (middle) and 1-WL color classes (right). More specifically, for GraphFSA, nodes cannot differentiate if they have 2 or 3 neighbors of a kind. In contrast, 1-WL can further distinguish the two nodes.}   
    \label{fig:expressivity}
\end{center}    
\end{figure}

\paragraph{Scalability.}
Scalability can be analyzed along two primary dimensions: the size of the graphs on which the cellular automata (CA) are executed, and the number of states that the learned CA can possess. Our approach demonstrates good scalability for larger graphs as it operates in linear time relative to the number of nodes and edges. However, scalability concerning the number of states is influenced by the chosen aggregation method. Specifically, with the counting aggregation, the number of possible aggregations becomes exponential in the number of states. It is, therefore, crucial to choose aggregations that lead to a smaller expansion when dealing with more states.

\begin{figure*}[t]
\begin{center}
    \centering
      \hspace{4em}
       \raisebox{0.33\height}{\includegraphics[width=0.43\textwidth]{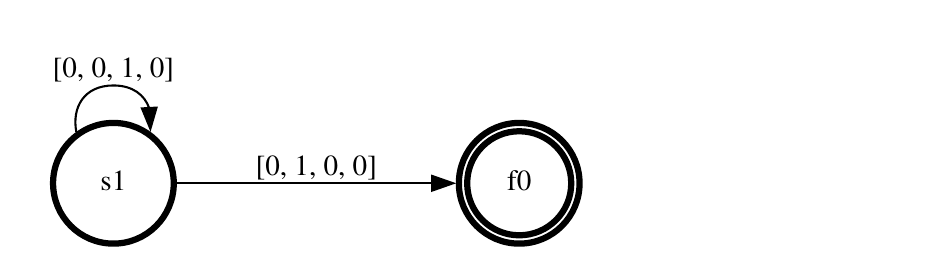}}
\hfill
    \includegraphics[width=0.45\textwidth]{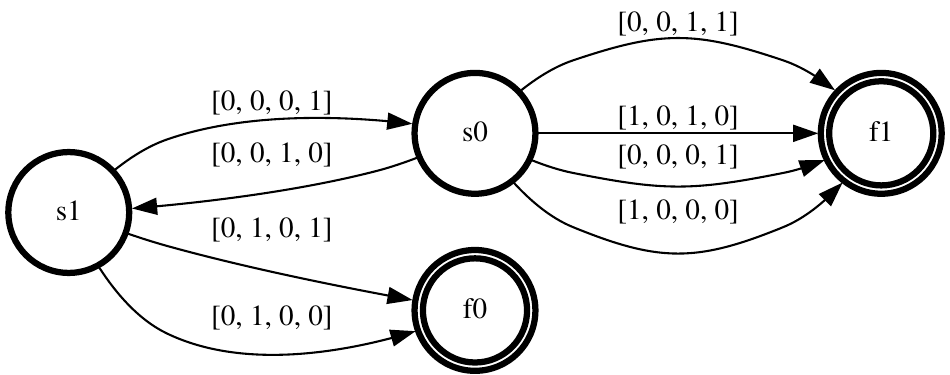}

\caption{Partial Visualization of a learned GraphFSA model for the Distance problem. The root node starts in state $s_1$ (left), whereas all other nodes start in state $s_0$ (right). The final states $f_0,f_1$ represent even and odd distances to the root, respectively. Aggregation values are presented as $[f_0, f_1, s_0, s_1]$ where we apply the counting aggregation with bounding parameter $b=1$. We can verify on the right that the nodes wait until they observe a final state in their neighborhood and then transition to the other final state. A full visualization of the same automaton can be found in the Appendix.}   
\label{fig:fsm-partial-example}

\end{center}    
\end{figure*}

\subsection{Expressiveness} 

While the execution of GraphFSA may bear resemblance to one round of the Weisfeiler-Lehman test, this does not hold in general. In the example Figure~\ref{fig:graphfsa-visualization}, they are the same because the chosen GraphFSA-instantiation is powerful enough to distinguish all occurrences of states appearing in the neighborhoods. More specifically, nodes can distinguish if they have \texttt{0} or \texttt{more than one} neighbor in a certain state (``1+ aggregation''), which is sufficient for the given graph. In general, the aggregation function $\mathcal{A}$ distinguishes GraphFSA from the standard WL-refinement notion as it restricts the observation of the neighborhood from the usual multiset observations (as these cannot be bounded in size). Consequently, GraphFSA is strictly less expressive than 1-WL, resulting in a tradeoff between using a more expressive aggregator and maintaining the simplicity or explainability of the resulting model, particularly as the state space grows exponentially. Figure~\ref{fig:expressivity} presents an example comparing GraphFSA's execution with a 2+ aggregator to WL refinement. Here, we can observe that GraphFSA is less expressive than 1-WL. However, for a large enough $b$, GraphFSA can match 1-WL on restricted graph classes of bounded degree. In conclusion, while the finite number of possible aggregations restricts GraphFSA, it also keeps the mechanism simple. Our evaluation demonstrates that this can improve generalization performance for a learned GraphFSA.

\subsection{Visualization and interpretability}
\label{sec:graph-fsa-visualization}

The GraphFSA framework offers inherent interpretability, facilitating visualization akin to a Finite State Automaton (FSA). This visualization approach allows us to easily explore the learned state transitions and the influence of neighboring nodes on state evolution. We outline two primary visualization techniques for GraphFSA:

\paragraph{Complete FSA visualization.} 
This method provides a comprehensive depiction of the entire FSA representation within the GraphFSA model, encompassing all states and transitions. Each node in the graph corresponds to a distinct state, while directed edges represent transitions between states based on transition values. An example is visualized in Figure ~\ref{fig:complex_visualization}.

\paragraph{Partial FSA visualization.} 
Tailored for targeted analysis of the GraphFSA model's behavior, this approach proves especially useful when dealing with large state spaces or focusing on specific behaviors. This method visualizes a selected starting state by evaluating the model across multiple problem instances and tracking all transitions that occur. Then, the visualization is constructed by exclusively displaying the states and transitions that have been used from the specified starting state.

Figure~\ref{fig:fsm-partial-example} illustrates the partial visualization of a learned automaton. The transitions in the diagram are determined by aggregating the information of the neighboring states. The structure of the transition value is $[f_0, f_1, s_0, s_1]$. For instance, the state vector $[0,0,1,0]$ indicates that the node in the graph has one or more neighbors in state $s_0$ while having no neighbors in other states.

\begin{figure*}[ht]
    \centering
    \includegraphics[width=0.17\textwidth]{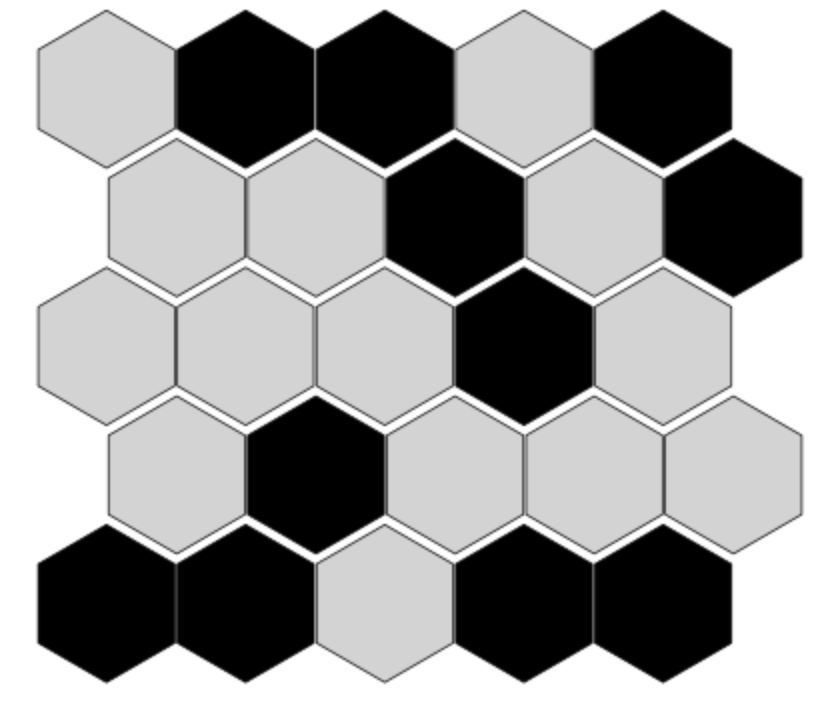}
    \hfill
    \includegraphics[width=0.17\textwidth]{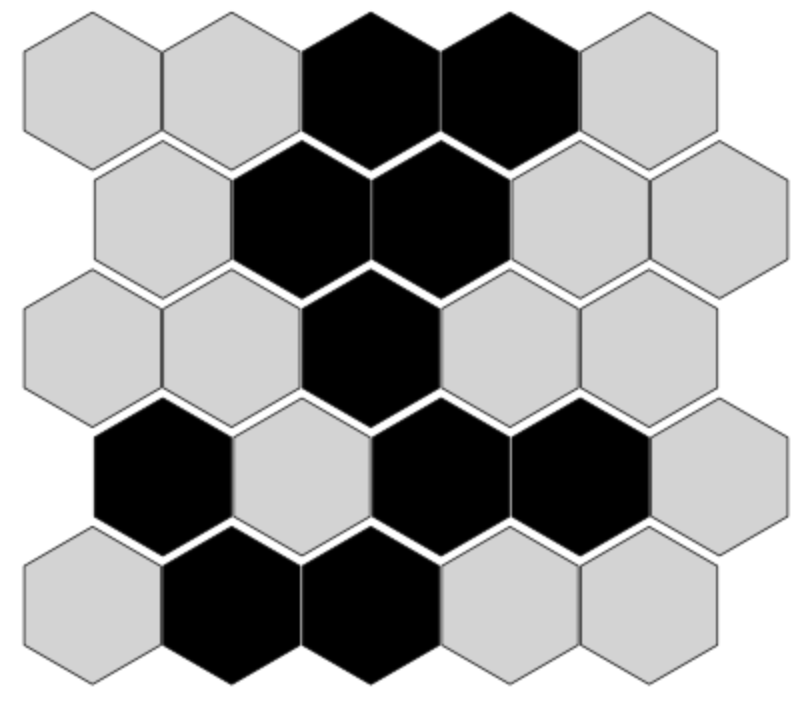}
    \hfill
    \includegraphics[width=0.55\textwidth]{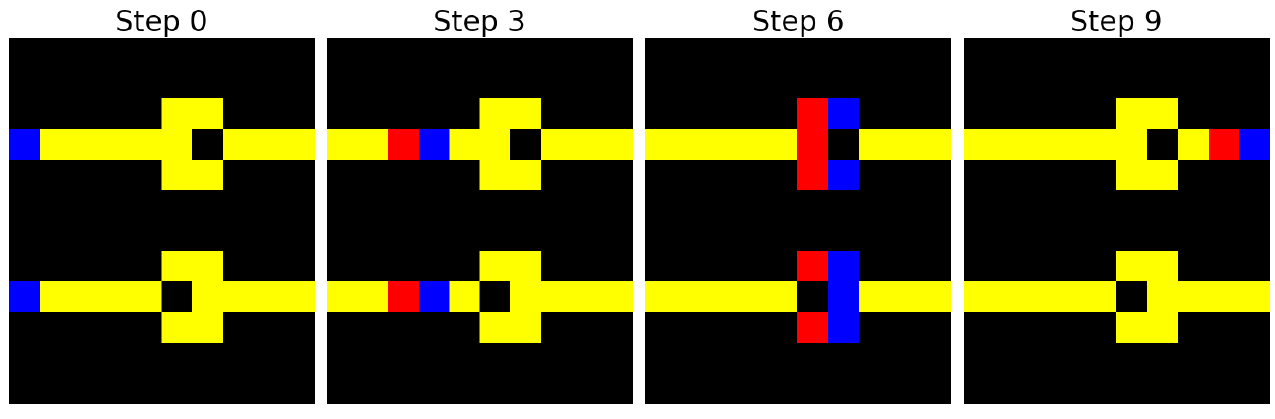}
    \label{fig:wireworld-dataset-illustration}
\caption{We study learning cellular automata as the simplest form of algorithmic behavior, which can lead to complex and intricate behavior patterns when assembled in a network. The left shows Conway's Game of Life, which we can learn on general topologies such as the hexagonal grid due to the graph representation. A WireWorld automata is depicted on the right, which can be used to simulate transistors and electrical circuits. It shows the electron head (blue) and tail (red) transitioning through different wires (yellow). 
}

\label{fig:hexagonal-gameoflife-dataset-illustration}

\end{figure*}

\subsection{GRAB: The GraphFSA dataset generator}
Accompanying the GraphFSA execution framework, which outlines the general interface and underlying principles shared amongst instances of GraphFSA, we provide GRAB, a flexible dataset generator to evaluate learned state automatons. Note that the introduced framework specifies the model structure but allows for different training methodologies that can be used to train specific model instances. The discrete nature of state transitions makes training these models a challenging task. This work proposes one possible way to train such models using input and output pairs. For alternative approaches, we deem it crucial to provide a systematic tool that can be used to develop new training strategies and further provide a principled way of thoroughly assessing the effectiveness across training methods.
Therefore, GRAB provides an evaluation framework for learning state automatons on graphs. GRAB generates the ground truth for a synthetic FSA, which can be used for learning and testing on various graphs. In particular, GRAB provides out-of-the-box support for several graph sizes and supports testing on multiple graph distributions.

In the dataset construction process, we first define the characteristics of the ground truth GraphFSA, specifying the number of states and configuring the number of initial and final states. Next, we generate a random Finite State Automaton (FSA) by selecting states for each transition value, adhering to the constraints of the final state set (final states cannot transition to other states). We then sample graphs from predefined distributions, offering testing on a diverse set of graph types such as trees, grids, fully connected graphs, regular graphs, and Erdos-Renyi graphs. The starting states for each node within these generated graphs are randomly initialized from the set $\mathcal{S}$. Finally, we apply the GraphFSA model to these input graphs to compute the outputs according to the defined state transitions. 
For the dataset generation, we emphasize evaluating generalization, which we believe to be one of the essential abilities of the GraphFSA framework. 
Therefore, we establish an extrapolation dataset to assess the model's capability beyond the graph sizes encountered during training. This dataset contains considerably larger graphs, representing a substantial deviation from the training data. However, if the correct automaton can be extracted during training, good generalization should be achievable despite the shift in the testing distribution. Input/output pairs for these larger graphs are produced by applying the same generated FSA to obtain the resulting states.

\section{Empirical evaluation}

The GraphFSA framework is designed to extract state automaton representations with discrete states and transitions to learn to solve the task at hand through an algorithmic representation. Our evaluation consists of two key components to showcase this.
In the first part, we focus on classic cellular automaton problems. These automatons serve as a foundational component of our study as they represent the simplest possible version of an algorithm. However, despite their simplicity, they can becomes increasingly powerful and exhibit complex behaviour when assembled together in a network. We successfully learn the underlying automaton rules, demonstrating the ability and advantage of the GraphFSA framework to capture simple algorithmic behaviour.\\

In the second part of our empirical study, we evaluate our models using our proposed graph automaton benchmark and a set of more challenging graph algorithms. GRAB enables a comprehensive investigation of GraphFSA's performance across a broad range of problems. We are particularly interested in the generalisation ability of the GraphFSA Framework. Therefore, we perform extrapolation experiments where we train on smaller graphs and subsequently test its performance on larger instances.
Finally, we use the GraphFSA framework to learn a selection of elementary graph algorithms, demonstrating the framework's capability and potential for algorithm learning.

\subsection{Training}
We take inspiration from prior research on training differentiable finite state machines~\citep{mordvintsev2022diff_fsm} and propose an adaption to train \model, which follows the GraphFSA framework, on general graphs.
To maintain differentiability, we relax the requirement for choosing a single next state for a transition in $\mathcal{T}$ during training and instead model it as a probability distribution over possible next states, resulting in a probabilistic automaton. 
We thus compute $P(X=m \ | \ \mathcal{M} \times \mathcal{Z})$ for $m \in \mathcal{M}$ and parameterize \model~ with a matrix $T'$ of size $\mathcal{M} \times \mathcal{Z} \times \mathcal{M}$ that holds the probabilities for all possible transitions and states. 
This means that the transition matrix $T'$ contains the probabilities of transitioning from a state $m_1$ to a state $m_2$ given a specific transition value.
To train the model with a given step number $t$, we execute \model~for $t$ rounds and compute the loss based on the ground-truth output states of the graph automaton we want to learn when executing the same number of steps $t$.
For example, in the case of node classification problems, we aim to predict each node's final state $f \in F$.  
This makes it possible to backpropagate through the transition matrix $T'$.
In addition, we apply an L2 loss to achieve the correct mapping between input and output states. 
Once the training of the probabilistic model is complete, we extract the decision matrix $T$ from $T'$ by discretizing the distribution $P$ over the next states to the most likely one.\\

Furthermore, we are interested in the model's ability to produce consistent and correct outputs, even as the number of iterations changes. We refer to this as \emph{Iteration stability}. The extrapolation capabilities of the GraphFSA model are achieved through the ability to adjust the number of iterations, representing the number of steps executed by the FSA. This flexibility is essential, especially for graphs with larger diameters, where information needs to propagate to all nodes. We can, therefore, train an FSA where we can choose a sufficiently large number of iterations to produce a correct output, and increasing this number of iterations still leads to a correct solution. We develop two approaches: 

\paragraph{Random iteration offset.} During training, we introduce random adjustments to the number of iterations with a small offset factor. This randomization ensures that the model can effectively handle the problem with varying numbers of iterations.

\paragraph{Final state loss.} \label{sec:final-state-loss}  One approach to incentivize final states not to switch to other states is to incorporate an additional loss term that penalizes leaving a final state. We consider the learned transition matrix probabilities and penalize the model if the probability of switching from a final state $f \in F$ to any other state $m \in \mathcal{M}, m \neq f$ is greater than 0 using an L1 loss. This approach encourages the model to stay in the final state, resulting in more stable predictions. Figure \ref{fig:fsm-partial-example} provides an example of a state automata trained with this additional loss.

\paragraph{Baseline models.}
The architecture that aligns most closely with our tasks is the Graph Neural Cellular Automata (GNCA)~\cite{grattarola2021learning}. While GNCAs are designed to learn rules from single-step supervision, we apply several rounds when training with multiple steps. 
Besides GNCAs, we use recurrent message-passing GNNs. In contrast to GNCA, we follow the typical encode-process-decode~\cite{Veli_kovi__2021}, which means that the GNN keeps intermediate hidden embeddings that can be decoded to make predictions. 
We use the architecture proposed by~\citet{AlgorithmicRecurrentGNN2022}, and observe that employing a GRU in the node update yields the best performance and generalization capabilities, in line with their results. This convolution is therefore used in all results in the paper. 
As the goal of a GraphFSA is to extract rules from the local neighborhood and update the node state purely based on this information, more advanced graph-learning architectures that go beyond this simple paradigm are not applicable in this setting. Our analysis mainly focuses on demonstrating the value of discrete state spaces, especially when generalizing to longer rollouts and larger graphs.

\subsection{Classical cellular automata problems}
As a first step towards learning algorithms, we focus on cellular automata probelms. We consider a variety of different settings such as 1D and 2D automata, these include more well known instances such as Wireworld or Game of Life, which despite their simple rules are known to be Turing complete. For a more detailed explanation of the datasets, we refer to the Appendix.

\paragraph{1D-Cellular Automata}
The 1D-Cellular Automata dataset consists of one-dimensional cellular automata systems, each defined by one of the possible $2^{2^3}$ rules. Each rule corresponds to a distinct mapping of a cell's state and its two neighbors to a new state.  

\paragraph{Game Of Life.} The GameOfLife dataset follows the rules defined by Conway's Game of Life~\citep{conway}. A cell becomes alive if it has exactly 3 live neighbors. If a cell has less than 2 or more than 4 live neighbors, it dies, otherwise it persists. Furthermore, we consider a 
a variation where cells are hexagonal instead of the traditional square grid, which we refer to as the Hexagonal Game of Life. A visual representation is depicted in Figure \ref{fig:hexagonal-gameoflife-dataset-illustration}.

\begin{figure}[t]
    \centering
    \includegraphics[width=0.4\linewidth]{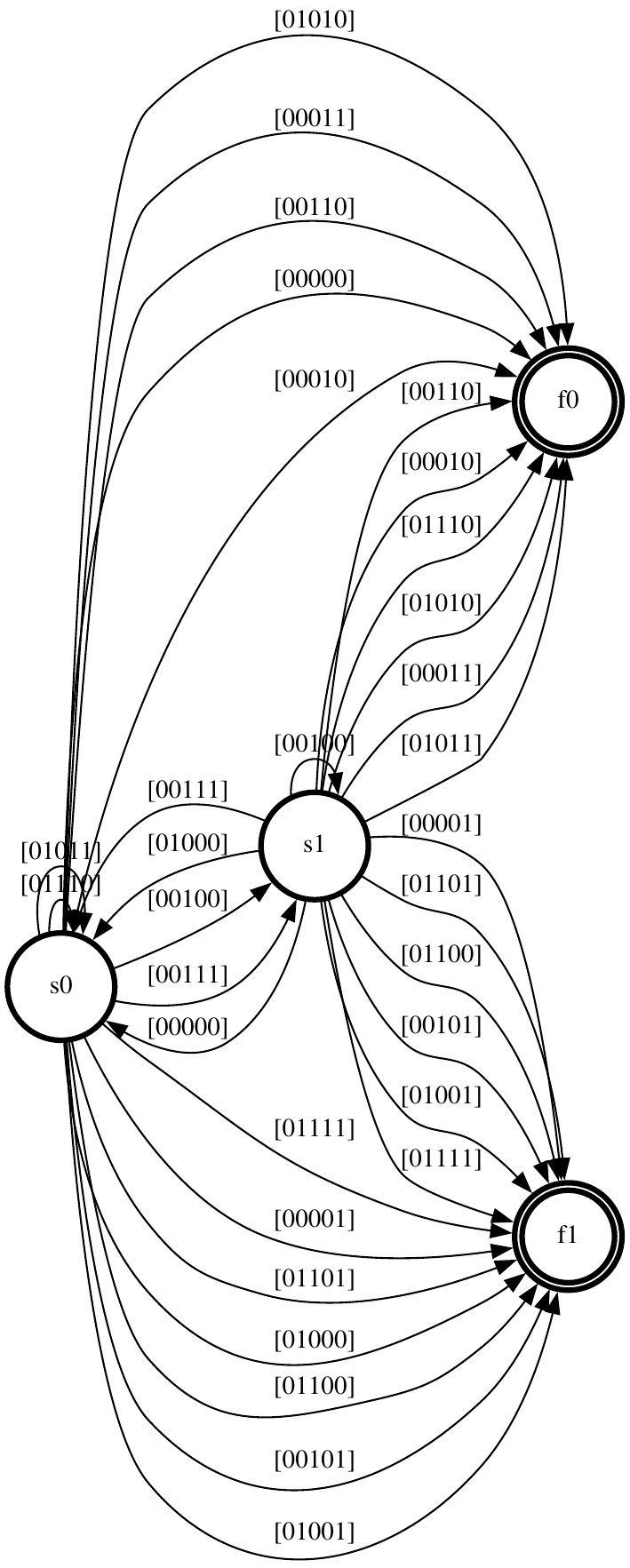}
    \caption{Complete Visualization of a learned GraphFSA model for the Distance problem. The root node starts in state $s_1$, whereas all other nodes start in state $s_0$. The final states $f_0,f_1$ represent even and odd distances to the root respectively. Aggregation values are presented as $[f_0, f_1, s_0, s_1]$ where we apply the counting aggregation with bounding parameter $b=1$.}
        \label{fig:complex_visualization}
\end{figure}

\paragraph{WireWorld.}
WireWorld ~\citep{wireworld} is an automaton designed to simulate the flow of an electron through a circuit. Cells can take one of four states: empty, electron head, electron tail, and conductor. Figure \ref{fig:hexagonal-gameoflife-dataset-illustration}
shows an initial grid starting state and the evolution of WireWorld over multiple time steps. 

\subsubsection{Results}
For Cellular Automata problems, we generate training data for each problem by initializing a 4x4 input grid or path of length 4 (for 1D problems) with random initial states and apply the problem's rules for a fixed number of steps to produce the corresponding output states. Our models are trained on these input-output pairs, using the same number of iterations as during dataset creation to ensure that we can extract the rule.
Additionally, we test on an extrapolation dataset consisting of larger 10x10 grids. This dataset is used to evaluate the model's ability to adapt to varying rule iterations with four different iteration numbers.

\paragraph{1D Cellular Automata.} 
On this dataset, \model~uses the ``positional neighbor-informed aggregation'' aggregation technique. Notably, our model can successfully learn the rules for one-step training data (t=1). For the complete results, we refer to the Appendix.

\begin{figure}[t]
    \centering
    \includegraphics[width=\linewidth]{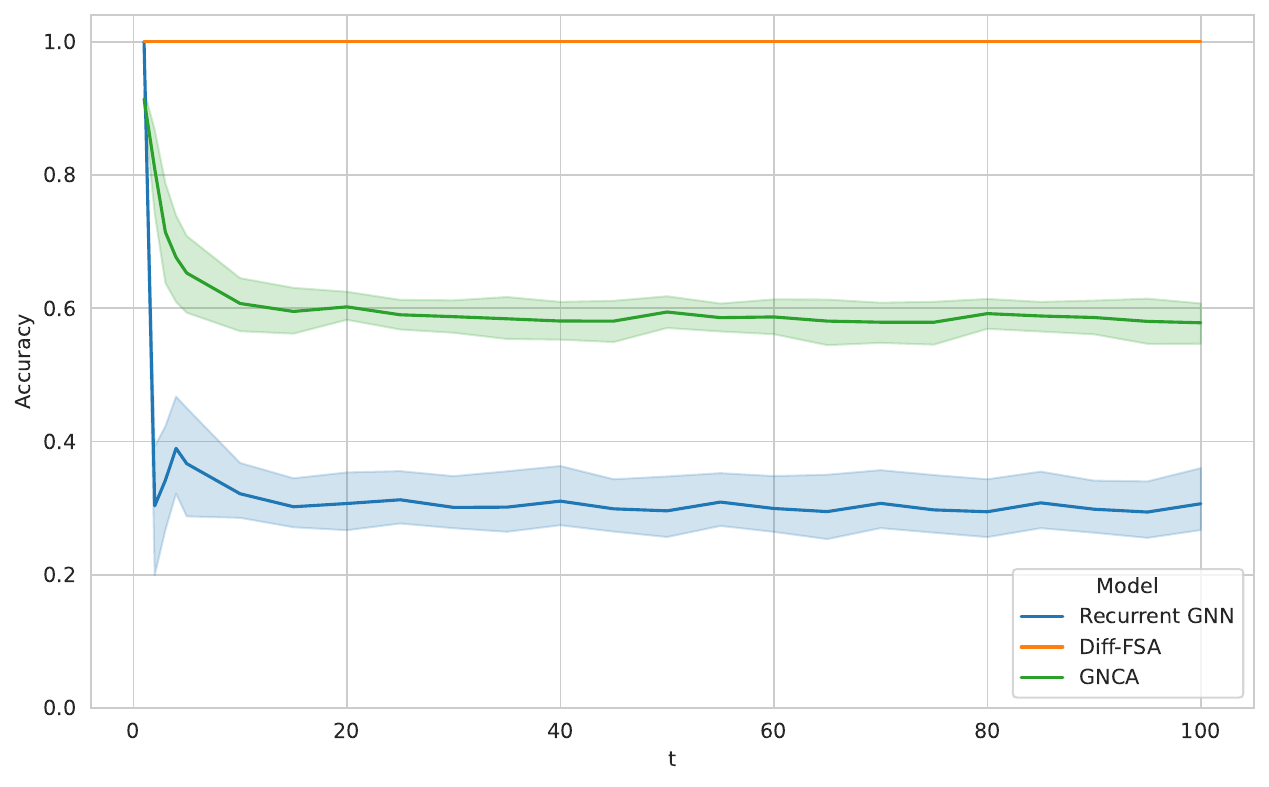}
    \caption{Mean accuracies for Wireworld on a regular grid when trained on 1 step and applied for $t$ steps during inference (for 10 seeds each). All models report high accuracy for $t=1$. However, performance deteriorates during inference when more steps are executed except for \model.}
        \label{fig:wire-results}
\end{figure}

\paragraph{2D Cellular Automata.} 
For 2D Cellular Automata (CA) problems, \model~ employs the counting aggregation. 
For Wireworld, we use a state space of four and a bounding parameter of three. We train all models to learn the one-step transition rules. During training, all models and baselines report high accuracy. We then investigate the generalization capabilities and especially the iteration stability of the learned models. We run the trained models on larger 10xs10 grids for more time steps t than during training. The results are depicted in Figure \ref{fig:wire-results}. Note that the recurrent GNN and GNCA both deteriorate outside the training distribution. However, \model~exhibits good iteration stability across the whole range of iterations.
For the Game of Life variations, we provide a detailed comparison in the Appendix.

\subsection{Evaluation with GRAB}
\begin{table*}[t]
\caption{
Evaluation of GraphFSA on synthetic data provided by GRAB. We report the accuracy and standard deviation over 10 runs. The underlying ground truth consists of an FSA using 4 states. We can test the in-distribution validation accuracy to see how well a model can fit the data. Moreover, we test extrapolation
to larger graphs to verify that the rules for underlying automata were successfully learned. Our \model~ models generally perform well across all scenarios. Note that the recurrent GNN performs well but lacks the interpretation and visualization of the learned mechanics as discrete state automata. 
}
    \label{tab:generator-result-main}
    \centering
    \resizebox{0.7\linewidth}{!}{
    \begin{tabular}{p{2.4cm}p{1.8cm}p{1.8cm}p{1.8cm}p{1.8cm}p{1.8cm}}
\toprule
Model & Val Acc & n=10 & n=20 & n=50 & n=100 \\
\midrule
GNCA & 0.38 $\pm$ 0.00 & 0.39 $\pm$ 0.00 & 0.40 $\pm$ 0.00 & 0.40 $\pm$ 0.00 & 0.39 $\pm$ 0.00 \\
Recurrent GNN & 1.00 $\pm$ 0.00 & 0.91 $\pm$ 0.10 & 0.85 $\pm$ 0.13 & 0.82 $\pm$ 0.16 & 0.81 $\pm$ 0.16 \\
\midrule
\model~(4 states)& 0.99 $\pm$ 0.02 & 0.97 $\pm$ 0.02 & 0.96 $\pm$ 0.02 & 0.95 $\pm$ 0.03 & 0.94 $\pm$ 0.03 \\
\model~(5 states) & 1.00 $\pm$ 0.01 & 0.98 $\pm$ 0.01 & 0.96 $\pm$ 0.02 & 0.95 $\pm$ 0.02 & 0.95 $\pm$ 0.02 \\
\model~(6 states) & 0.99 $\pm$ 0.01 & 0.98 $\pm$ 0.01 & 0.95 $\pm$ 0.02 & 0.94 $\pm$ 0.02 & 0.94 $\pm$ 0.02 \\
\bottomrule
\end{tabular}}
\end{table*}

We use GRAB to create diverse datasets to benchmark the different models. These datasets encompass different graph types and sizes and generate a synthetic state automaton to generate the ground truth. This allows us to precisely control the setup by adjusting the number of states of the ground truth automaton. For each specific dataset configuration, we train 10 different model instances. Extended results with more runs can be found in the Appendix.

Table~\ref{tab:generator-result-main} shows the results for a dataset generated by GRAB. The dataset consists of tree graphs with a ground truth FSA consisting of 4 states. Generally, we can observe that both \model instances of the GraphFSA framework perform well, even if the number of states does not match exactly. The GNCA model performs poorly in this setting, as it struggles to learn the rules when they are applied for multiple steps. The recurrent GNN baseline performs well on in-distribution graphs, but struggles to generalize when the graph sizes are increased. Note that the recurrent GNN operates on continuous states, in contrast to the discrete states used by~\model.
Moreover, the experiments were conducted with a varying number of states for \model. While the ground-truth solution is modeled with 4 states, we want to investigate whether increasing this number for the learned automaton leads to different results. As we can observe, the accuracies are consistent for all sizes, indicating that more states can be used for training, which is especially useful when the underlying ground-truth automaton is not known beforehand. 

\subsection{Graph algorithms}
\begin{table}
    \centering
    \caption{
   Evaluation of learning graph algorithms on the Distance dataset. We report the accuracy and standard deviation over 10 runs. All models are trained on graphs of size at most 10 and then tested for extrapolation on larger graph sizes $n$. Our \model~ outperforms other discrete transition-based models while matching the recurrent GNN baseline.}
    \resizebox{\linewidth}{!}{
    \begin{tabular}{lllll}
\toprule
Model & n = 10 & n = 20 & n = 50 & n = 100 \\
\midrule
GNCA & 0.50 $\pm$ 0.00 & 0.50 $\pm$ 0.00 & 0.51 $\pm$ 0.00 & 0.50 $\pm$ 0.00 \\
Recurrent GNN & 1.00 $\pm$ 0.00 & 1.00 $\pm$ 0.00 & 1.00 $\pm$ 0.00 & 1.00 $\pm$ 0.00 \\
\midrule
\model & 1.00 $\pm$ 0.00 & 1.00 $\pm$ 0.00 & 1.00 $\pm$ 0.00 & 1.00 $\pm$ 0.00 \\
\bottomrule
\end{tabular}
}
    \label{tab:graph-algorithms}
\end{table}

To showcase the potential of the GraphFSA framework for algorithm learning, we further evaluate it on more challenging graph algorithm problems. We use the same graph algorithms as~\citet{AlgorithmicRecurrentGNN2022}, which, among other setups, includes a simplified distance computation and path finding. Detailed descriptions of these datasets can be found in the Appendix.
Similar to the data generated by GRAB, our training process incorporates a random number of iterations, with the sole condition that the number of steps is sufficiently large to ensure complete information propagation across the graph. For problems involving small training graphs, we train the \model~ model with approximately ten iterations and apply the additive final state loss to ensure iteration stability and establish distinct starting and final states.\\

The main aim is to validate that our proposed model is capable of learning more challenging algorithmic problems. It consistently performs better than GNCA. Compared to the recurrent GNN, it does not perform as well across all selected problems, but can learn correct solutions, as seen in the Distance task illustrated in Table~\ref{tab:graph-algorithms}. However, recall that ~\model operates on discrete states, making it much more challenging to learn. On the other hand, we can extract the learned solution and analyse the learned mechanics.
A visualization of such a learned model is depicted in Figure~\ref{fig:fsm-partial-example} and Figure~\ref{fig:complex_visualization}.
This has the advantage over the recurrent baseline in that the learned mechanics can be interpreted, and the algorithmic behavior can be explained and verified. 
\section{Limitations and future work}
The main advantage of the GraphFSA framework is its use of discrete states. This allows us to interpret the learned mechanics through the lens of a state automaton. Moreover, it can perform well in scenarios where the underlying rules of a problem can be modeled with discrete transitions while requiring that inputs and outputs can be represented as discrete states. 
This is the case for several algorithmic settings but limits the model's applicability to arbitrary graph learning tasks. To broaden the applicability of the GraphFSA framework, future work could investigate methods to map continuous input values to discrete states, potentially with a trainable module. 
Another aspect to investigate is the study of more finite aggregations for GraphFSA. These can heavily influence a model's expressivity or align its execution to a specific task.
Moreover, \model~only represents one possible approach to train models within the GraphFSA framework. Training models that yield discrete transitions remains challenging and could further improve performance and effectiveness.

\section{Conclusion}

We present GraphFSA, an execution framework that extends finite state automata by leveraging discrete state spaces on graphs. Our research demonstrates the potential of GraphFSA for algorithm learning on graphs due to its ability to simulate discrete decision-making processes. Additionally, we introduce GRAB, a benchmark designed to test and compare methods compatible with the GraphFSA framework across a variety of graph distributions and sizes.
Our evaluation shows that \model can effectively learn rules for classical cellular automata problems, such as the Game of Life, producing structured and interpretable representations in the form of finite state automata. While this approach is intentionally restrictive, it simplifies complexity and aligns the model with the task at hand.
We further validate our methodology on a range of synthetic state automaton problems and complex algorithmic datasets. Notably, the discrete state space within GraphFSA enables \model to exhibit robust generalization capabilities.

\newpage

\bibliography{mybibfile}

\newpage
\appendix

\section{Extended dataset descriptions}
\subsection{Cellular Automata}
\label{sec:extended-ca-dataset}
As a first interesting step towards learning algorithms, we focus on cellular automata probelms. We consider a variety of different settings such as 1D and 2D automata, these include more well known instances such as Wireworld or Game of Life, which despite its simple behaviour is known to be Turing complete.

\begin{figure}[H]
    \begin{center}
        \includegraphics[scale=0.23]{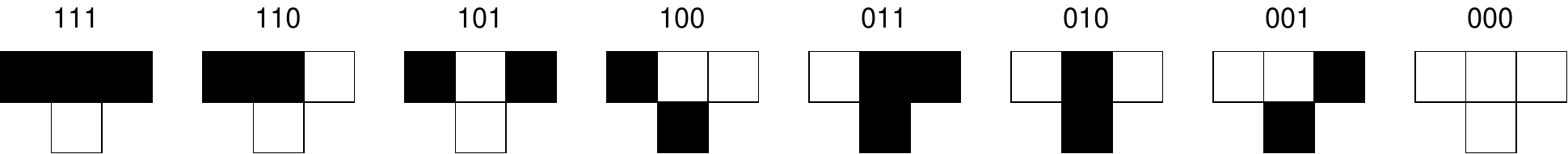}
        \caption{Visualization of 1D CA rule 30 - top row shows different combinations (left neighbor state, current cell state, right neighbor state), bottom row center shows new state value for the cell}   
                \label{fig:1dca-illustration}

    \end{center}    
\end{figure}

\paragraph{1D-Cellular Automata}
The 1D-Cellular Automata dataset consists of one-dimensional cellular automata systems, each defined by one of the possible 256 rules. Each rule corresponds to a distinct mapping of a cell's state and its two neighbors to a new state. Figure \ref{fig:1dca-illustration} provides a graphical illustration of one such automata. Note, that we need todistinguish between the left and right neighbors in order to capture all rules in GraphFSA.

\paragraph{Game Of Life} The GameOfLife dataset captures the essence of Conway's Game of Life that progresses based on its initial state and a set of simple rules. 

The progression of Conway's Game of Life is dictated by a set of simple rules applied to each cell in the grid, considering its neighbors. Using the Moore neighborhood, which includes all eight surrounding cells, these rules are as follows:
\begin{enumerate}
\item Birth: A cell that is dead will become alive if it has exactly three living neighbors.
\item Survival: A cell that is alive will remain alive if it has two or three living neighbors.
\item Death:
\begin{enumerate}
\item \textit{Loneliness}: A cell that is alive and has fewer than two living neighbors will die, mimicking underpopulation.
\item \textit{Overpopulation}: A cell that is alive and has more than three living neighbors will die, representing overpopulation.
\end{enumerate}
\end{enumerate}

\paragraph{Toroidal vs. non-toroidal}
For this dataset, we consider both toroidal and non-toroidal variations: 

\begin{enumerate}
    \item \textit{Toroidal}: In the toroidal variation, the board's edges wrap around, creating a continuous, closed surface. This means cells on the edge have neighbors on the opposite edge.
    \item \textit{Non-Toroidal}: In the standard, non-toroidal variation, cells on the board's edge only consider neighbors within the boundary.
\end{enumerate}

Our dataset consists of input/output pairs where we randomly initialize the grid and then apply the Game of Life rules for a fixed number of steps. We represent this dataset through grid graphs and use the Moore neighborhood.

\paragraph{Hexagonal Game Of Life} 
The Hexagonal Game Of Life introduces a variation where cells are hexagonal as opposed to the traditional square grid. This change in cell structure offers a fresh set of neighbour cells, which can lead to distinct patterns and evolutions. A visual representation can be found in Figure \ref{fig:hexagonal-gameoflife-dataset-illustration}.

\paragraph{WireWorld}
The WireWorld dataset revolves around the cellular automaton 'WireWorld' where cells can take one of four states: empty, electron head, electron tail, and conductor. It's especially renowned for its capability to implement digital circuitry. In the dataset, we observe the evolution of a given cellular configuration over specified iterations.

\subsection{Graph algorithms}
\label{sec:extended-graph-algorithms}

Taking inspiration from the datasets utilized by \citet{AlgorithmicRecurrentGNN2022}, we create datasets for various classical graph problems. To ensure versatility and scalability, we generate new graphs for each problem instance and compute the corresponding ground truth during dataset creation. This approach enables us to construct datasets that not only encompass graphs of specific sizes but also facilitate evaluation on larger extrapolation datasets. We explore the following graph problems that helps us to explore different capabilities of our model.

\subsubsection{Distance}
The distance problem involves determining whether each node in the graph has an even or odd distance from the root node. To formulate this problem, we define input values for each problem instance, representing each node's state in the graph. Among these nodes, one is designated as the root, while the others are marked as non-root inputs. The output is assigned a binary value (0 or 1) for each node based on $\text{distance} \mod 2$ from the root, where ``distance'' represents the length of the shortest path between the root and a node.

\subsubsection{RootValue} In the root value problem, we want to propagate a value from the root throughout a path graph. One node in the graph is assigned a root label and a binary value (0,1). The objective is to propagate this binary value from the root across the entire graph.

\subsubsection{PathFinding}
The PathFinding problem determines whether a given node lies on the path between two marked nodes within a tree. The dataset comprises different trees, and two nodes are explicitly marked as relevant nodes within each tree. The objective is to predict whether a specific node in the tree, which is not one of the labeled nodes, lies on the path connecting these two marked nodes.

\subsubsection{PrefixSum}
The PrefixSum Dataset involves paths represented as sequences of nodes where each has an initial binary feature (either 1 or 0). The task is to predict the sum of all initial features to the right of each node in the path, considering modulo two arithmetic.

\section{Additional model evaluation}

\subsection{1D cellular automata}
We evaluate the models on 1D Cellular Automata evaluation for the different baselines and consider a larger graph during and multiple timesteps $t$ for our evaluation. We report the results in Table \ref{tab:1d-1step} for models that were trained on $t=1$ and in Table \ref{tab:1d-2step} for models trained on $t=2$.
\begin{table}
    \centering
\caption{1D Cellular Automata evaluation for the different baselines, we consider a path of size $10$ for the extrapolation data, and $t$ indicates the number of times the CA rule has been applied. We report accuracy and all models were trained for $t=1$. }    \resizebox{\linewidth}{!}{
\begin{tabular}{llllllll}
\toprule
Model & t = 1 & t = 2 & t = 5 & t = 10 & t = 20 & t = 50 & t = 100 \\\midrule
\midrule
GNCA & 0.75 $\pm$ 0.00 & 0.66 $\pm$ 0.05 & 0.71 $\pm$ 0.08 & 0.79 $\pm$ 0.17 & 0.78 $\pm$ 0.17 & 0.77 $\pm$ 0.18 & 0.77 $\pm$ 0.18 \\
Recurrent GNN & 0.75 $\pm$ 0.00 & 0.50 $\pm$ 0.08 & 0.47 $\pm$ 0.17 & 0.47 $\pm$ 0.29 & 0.47 $\pm$ 0.30 & 0.47 $\pm$ 0.30 & 0.47 $\pm$ 0.30 \\
\midrule
\model & 1.00 $\pm$ 0.00 & 1.00 $\pm$ 0.00 & 1.00 $\pm$ 0.00 & 1.00 $\pm$ 0.00 & 1.00 $\pm$ 0.00 & 1.00 $\pm$ 0.00 & 1.00 $\pm$ 0.00 \\
\bottomrule

\end{tabular}
}
    \label{tab:1d-1step}
\end{table}

\begin{table}
    \centering
\caption{1D Cellular Automata evaluation for the different baselines, we consider a path of size $10$ for the extrapolation data, and $t$ indicates the number of times the CA rule has been applied. We report accuracy and all models were trained for $t=2$. }    \resizebox{\linewidth}{!}{
\begin{tabular}{llllllll}
\toprule
Model & t = 1 & t = 2 & t = 5 & t = 10 & t = 20 & t = 50 & t = 100 \\\midrule
GNCA & 0.49 $\pm$ 0.11 & 0.73 $\pm$ 0.03 & 0.70 $\pm$ 0.13 & 0.86 $\pm$ 0.16 & 0.86 $\pm$ 0.16 & 0.86 $\pm$ 0.16 & 0.86 $\pm$ 0.16 \\
Recurrent GNN & 0.64 $\pm$ 0.15 & 0.79 $\pm$ 0.00 & 0.58 $\pm$ 0.13 & 0.66 $\pm$ 0.26 & 0.67 $\pm$ 0.29 & 0.67 $\pm$ 0.30 & 0.68 $\pm$ 0.30 \\
\midrule
\model & 1.00 $\pm$ 0.00 & 1.00 $\pm$ 0.00 & 1.00 $\pm$ 0.00 & 1.00 $\pm$ 0.00 & 1.00 $\pm$ 0.00 & 1.00 $\pm$ 0.00 & 1.00 $\pm$ 0.00 \\
\bottomrule
\end{tabular}
}
    \label{tab:1d-2step}
\end{table}

\subsection{2D cellular automata}

\begin{figure}
    \centering
    \includegraphics[width=\linewidth]{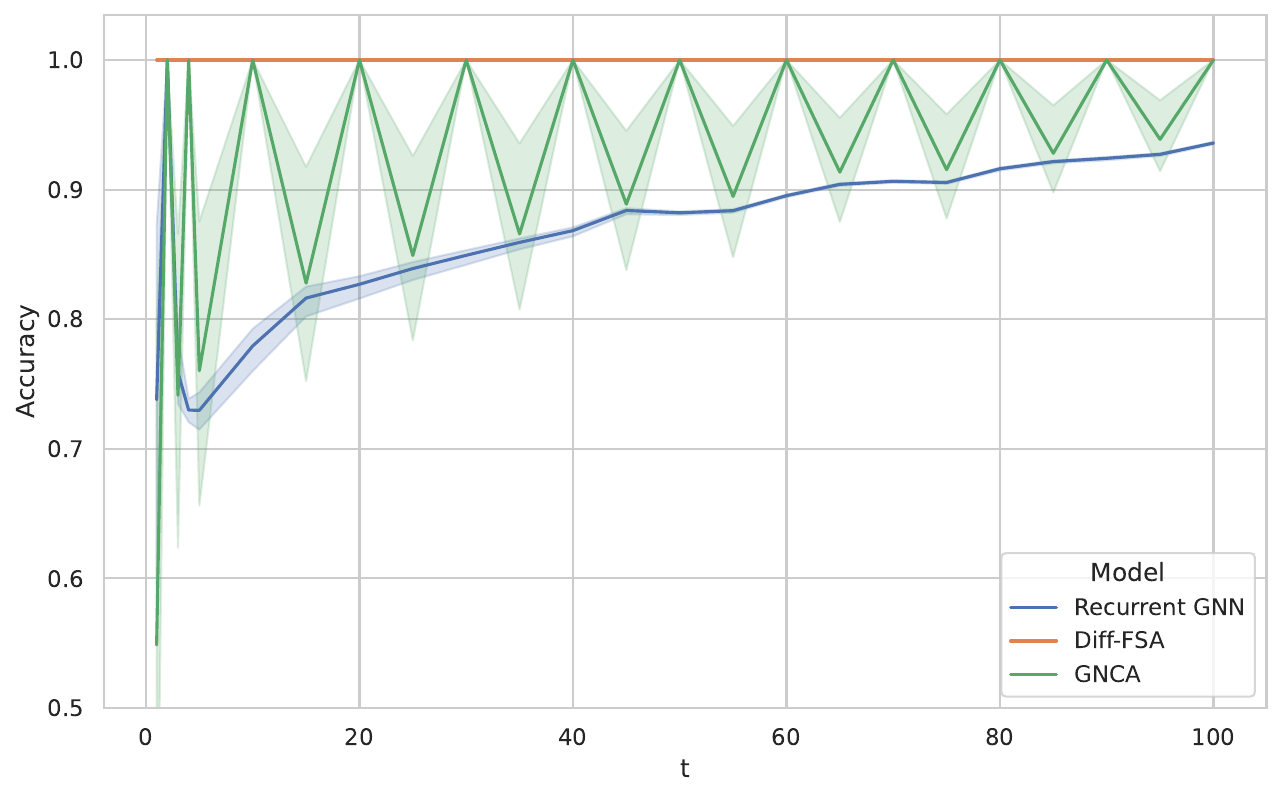}
    \caption{Mean accuracies for Game of Life on a regular grid when learned with 2 steps and applied for $t$ steps during inference (for 10 seeds each). All models report high accuracy for the training setup. However, performance deteriorates during inference when more steps are executed except for \model.}
        \label{fig:gol-results}
\end{figure} 

\begin{figure}
    \centering
    \includegraphics[width=\linewidth]{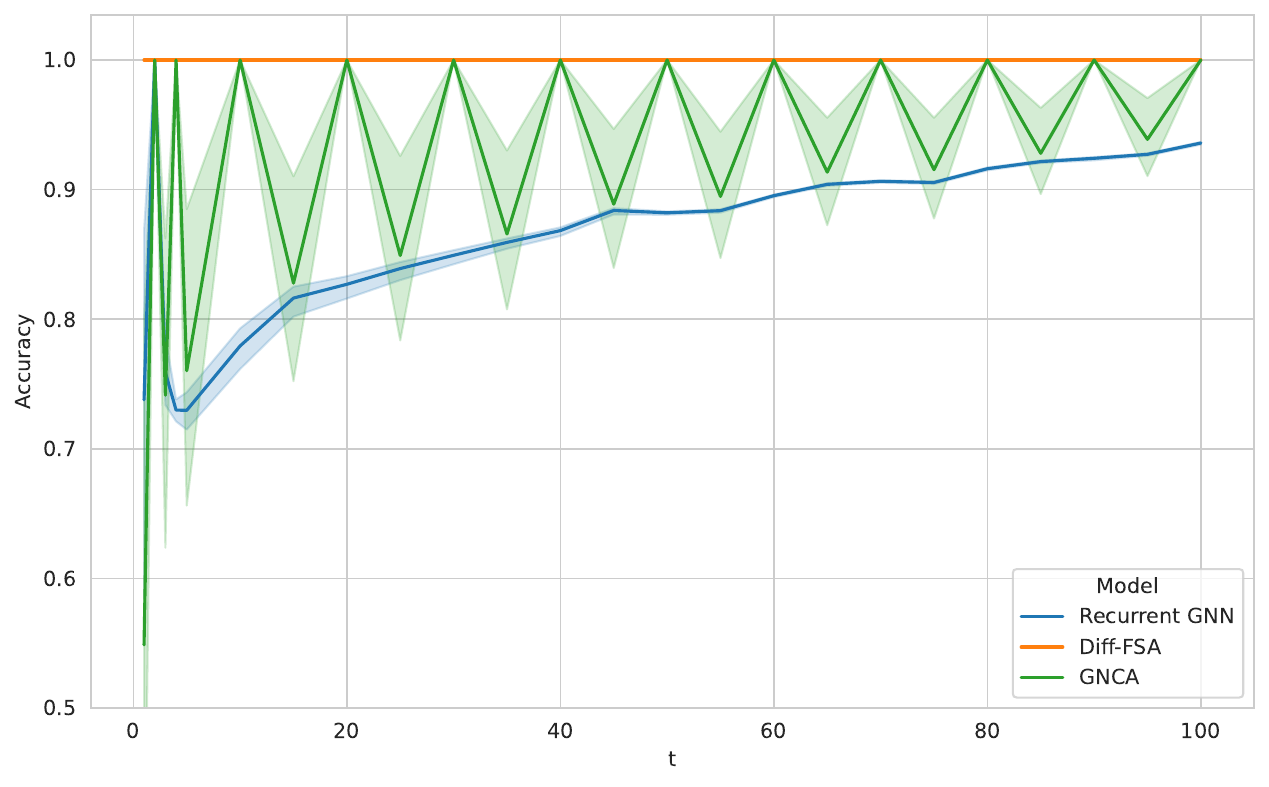}
    \caption{Mean accuracies for Game of Life on a hexagonal grid when learned with 2 steps and applied for $t$ steps during inference (for 10 seeds each). All models report high accuracy for the training setup. However, performance deteriorates during inference when more steps are executed except for \model.}
        \label{fig:gol-hex-results}
\end{figure}

\paragraph{Game of Life results.}Specifically, we use a state space of two for Game of Life and a bounding parameter of five. We let all models execute 2 steps in order to learn the appropriate transition rules after two iterations for Game of Life. Further, we investigate the generalization capabilities and especially the iteration stability of the learned models. For this, we consider all models which achieve perfect training accuracy and let them run on larger 10 node paths for more timesteps t than during training. The results are depicted in Figure \ref{fig:gol-results} for the regular Game of Life and in Figure \ref{fig:gol-hex-results} for the hexagonal variant. Note, that the recurrent GNN deteriorates outside the training distribution. Similarly, the GNCA baseline struggles to generalize as it has not learned the correct intermediate transitions for odd number of rules. The \model~ on the other hand exhibits good iteration stability across the whole range of iterations.

\subsection{Dataset generator}
\label{sec:experiments-dataset}
We perform further evaluations on more datasets generated by our dataset generator GRAB. The groudturth automaton consists of 4 states and we test multiple \model with 4, 5, and 6 states as well as the baselines over 10 runs. We report the achieved accuracies for the different experiments in Table \ref{tab:generator-result}.

\begin{table*}
\caption{
Evaluation of GraphFSA on synthetic data provided by GRAB. We report the accuracy and standard deviation over 10 runs. The underlying ground truth consists of an FSA using 4 states. We can test the in-distribution validation accuracy to see how well a model can fit the data. Moreover, we test extrapolation
to larger graphs to verify that the rules for underlying automata were successfully learned. Our \model~ models generally perform well across all scenarios. Note that the recurrent GNN performs well but lacks the interpretation and visualization of the learned mechanics as discrete state automata. 
}
    \label{tab:generator-result}
    \centering
    \resizebox{0.95\textwidth}{!}{
    \begin{tabular}{p{2.4cm}p{1.8cm}p{1.8cm}p{1.8cm}p{1.8cm}p{1.8cm}}
\toprule
Model & Val Acc & n=10 & n=20 & n=50 & n=100 \\
\midrule
\multicolumn{6}{c}{ Experiment 1 } \\ \midrule
GNCA & 0.38 $\pm$ 0.00 & 0.39 $\pm$ 0.00 & 0.40 $\pm$ 0.00 & 0.40 $\pm$ 0.00 & 0.39 $\pm$ 0.00 \\
Recurrent GNN & 1.00 $\pm$ 0.00 & 0.91 $\pm$ 0.10 & 0.85 $\pm$ 0.13 & 0.82 $\pm$ 0.16 & 0.81 $\pm$ 0.16 \\
\midrule
\model~(4 states)& 0.99 $\pm$ 0.02 & 0.97 $\pm$ 0.02 & 0.96 $\pm$ 0.02 & 0.95 $\pm$ 0.03 & 0.94 $\pm$ 0.03 \\
\model~(5 states) & 1.00 $\pm$ 0.01 & 0.98 $\pm$ 0.01 & 0.96 $\pm$ 0.02 & 0.95 $\pm$ 0.02 & 0.95 $\pm$ 0.02 \\
\model~(6 states) & 0.99 $\pm$ 0.01 & 0.98 $\pm$ 0.01 & 0.95 $\pm$ 0.02 & 0.94 $\pm$ 0.02 & 0.94 $\pm$ 0.02 \\
\midrule
\multicolumn{6}{c}{ Experiment 2 } \\ \midrule
GNCA & 0.66 $\pm$ 0.00 & 0.67 $\pm$ 0.00 & 0.66 $\pm$ 0.00 & 0.65 $\pm$ 0.00 & 0.65 $\pm$ 0.00 \\
Recurrent GNN & 1.00 $\pm$ 0.00 & 0.84 $\pm$ 0.02 & 0.82 $\pm$ 0.02 & 0.81 $\pm$ 0.02 & 0.80 $\pm$ 0.02 \\
\midrule
\model~(4 states) & 1.00 $\pm$ 0.00 & 0.94 $\pm$ 0.01 & 0.93 $\pm$ 0.02 & 0.92 $\pm$ 0.02 & 0.91 $\pm$ 0.02 \\
\model~(5 states) & 1.00 $\pm$ 0.00 & 0.93 $\pm$ 0.02 & 0.92 $\pm$ 0.02 & 0.91 $\pm$ 0.02 & 0.90 $\pm$ 0.02 \\
\model~(6 states) & 1.00 $\pm$ 0.00 & 0.94 $\pm$ 0.02 & 0.93 $\pm$ 0.02 & 0.92 $\pm$ 0.02 & 0.91 $\pm$ 0.02 \\
\midrule
\multicolumn{6}{c}{ Experiment 3 } \\ \midrule
GNCA & 0.92 $\pm$ 0.00 & 0.95 $\pm$ 0.00 & 0.93 $\pm$ 0.00 & 0.93 $\pm$ 0.00 & 0.93 $\pm$ 0.00 \\
Recurrent GNN & 0.98 $\pm$ 0.03 & 0.98 $\pm$ 0.02 & 0.96 $\pm$ 0.03 & 0.95 $\pm$ 0.03 & 0.95 $\pm$ 0.03 \\
\midrule
\model~(4 states) & 0.92 $\pm$ 0.00 & 0.95 $\pm$ 0.00 & 0.93 $\pm$ 0.00 & 0.93 $\pm$ 0.00 & 0.93 $\pm$ 0.00 \\
\model~(5 states) & 0.91 $\pm$ 0.02 & 0.94 $\pm$ 0.03 & 0.92 $\pm$ 0.01 & 0.93 $\pm$ 0.01 & 0.93 $\pm$ 0.01 \\
\model~(6 states) & 0.91 $\pm$ 0.02 & 0.94 $\pm$ 0.03 & 0.92 $\pm$ 0.01 & 0.93 $\pm$ 0.01 & 0.93 $\pm$ 0.01 \\
\midrule
\multicolumn{6}{c}{ Experiment 4 } \\ \midrule
GNCA & 0.56 $\pm$ 0.00 & 0.64 $\pm$ 0.00 & 0.63 $\pm$ 0.00 & 0.64 $\pm$ 0.00 & 0.63 $\pm$ 0.00 \\
Recurrent GNN & 1.00 $\pm$ 0.00 & 0.98 $\pm$ 0.05 & 0.97 $\pm$ 0.05 & 0.97 $\pm$ 0.05 & 0.97 $\pm$ 0.05 \\
\midrule
\model~(4 states) & 1.00 $\pm$ 0.00 & 1.00 $\pm$ 0.00 & 1.00 $\pm$ 0.00 & 1.00 $\pm$ 0.00 & 1.00 $\pm$ 0.00 \\
\model~(5 states) & 1.00 $\pm$ 0.00 & 1.00 $\pm$ 0.00 & 1.00 $\pm$ 0.00 & 1.00 $\pm$ 0.00 & 1.00 $\pm$ 0.00 \\
\model~(6 states) & 1.00 $\pm$ 0.00 & 1.00 $\pm$ 0.00 & 1.00 $\pm$ 0.00 & 1.00 $\pm$ 0.00 & 1.00 $\pm$ 0.00 \\
\midrule
\multicolumn{6}{c}{ Experiment 5 } \\ \midrule
GNCA & 0.54 $\pm$ 0.00 & 0.54 $\pm$ 0.00 & 0.55 $\pm$ 0.00 & 0.54 $\pm$ 0.00 & 0.55 $\pm$ 0.00 \\
Recurrent GNN & 1.00 $\pm$ 0.00 & 0.97 $\pm$ 0.03 & 0.94 $\pm$ 0.03 & 0.93 $\pm$ 0.03 & 0.93 $\pm$ 0.04 \\
\midrule
\model~(4 states) & 0.98 $\pm$ 0.00 & 0.96 $\pm$ 0.00 & 0.95 $\pm$ 0.00 & 0.93 $\pm$ 0.00 & 0.93 $\pm$ 0.00 \\
\model~(5 states) & 0.98 $\pm$ 0.00 & 0.97 $\pm$ 0.00 & 0.95 $\pm$ 0.00 & 0.94 $\pm$ 0.00 & 0.94 $\pm$ 0.00 \\
\model~(6 states) & 0.98 $\pm$ 0.00 & 0.96 $\pm$ 0.01 & 0.95 $\pm$ 0.01 & 0.94 $\pm$ 0.01 & 0.94 $\pm$ 0.00 \\

\bottomrule
\end{tabular}}
\end{table*}

\subsection{Algorithms}
We evaluate the GraphFSA framework and in particular our \model and the baselines on more challenging algorithmic datasets. We evaluate on the four datasets Distance, PrefixSum, PathFinding and RootValue. The report the results in the tables: \ref{tab:rootvalue-results}, \ref{tab:prefixsum-result} and \ref{tab:pathfinding-result}.
\label{sec:algorithm-extended-evaluation}
\begin{table}[H]
    \centering
    \caption{Evaluation of learning graph algorithms on the RootValue dataset. We report the accuracy and standard deviation over 10 runs. All models are trained on graphs of size at most 10 and then tested for extrapolation on larger graph sizes $n$.}
    \resizebox{\columnwidth}{!}{
    \begin{tabular}{ p{3.6cm}p{2cm}p{2cm}p{2cm}p{2cm}  }
\toprule
Model & n = 10 & n = 20 & n = 50 & n = 100 \\
\midrule
GNCA & 0.50 $\pm$ 0.00 & 0.50 $\pm$ 0.00 & 0.50 $\pm$ 0.00 & 0.50 $\pm$ 0.00 \\
Recurrent GNN & 1.00 $\pm$ 0.00 & 1.00 $\pm$ 0.00 & 0.92 $\pm$ 0.16 & 0.91 $\pm$ 0.16 \\
\midrule
\model & 1.00 $\pm$ 0.02 & 0.99 $\pm$ 0.02 & 0.99 $\pm$ 0.03 & 0.99 $\pm$ 0.04 \\
\bottomrule
\end{tabular}
}

\label{tab:rootvalue-results}

\end{table}

\begin{table}[H]
    \centering
    \caption{Evaluation of learning graph algorithms on the PathFinding dataset. We report the accuracy and standard deviation over 10 runs. All models are trained on graphs of size at most 10 and then tested for extrapolation on larger graph sizes $n$.}
    \resizebox{\columnwidth}{!}{
    \begin{tabular}{lllll}
\toprule
Model & n = 10 & n = 20 & n = 50 & n = 100 \\
\midrule
GNCA & 0.58 $\pm$ 0.00 & 0.73 $\pm$ 0.00 & 0.83 $\pm$ 0.00 & 0.89 $\pm$ 0.00 \\
Recurrent GNN & 0.97 $\pm$ 0.02 & 0.90 $\pm$ 0.03 & 0.88 $\pm$ 0.04 & 0.89 $\pm$ 0.09 \\
\midrule
\model & 0.85 $\pm$ 0.00 & 0.85 $\pm$ 0.00 & 0.87 $\pm$ 0.00 & 0.91 $\pm$ 0.00 \\
\bottomrule
\end{tabular}
}
    \label{tab:pathfinding-result}
\end{table}

\begin{table}[H]
    \centering
    \caption{Evaluation of learning graph algorithms on the PrefixSum dataset. We report the accuracy and standard deviation over 10 runs. All models are trained on graphs of size at most 10 and then tested for extrapolation on larger graph sizes $n$. }
    \resizebox{\columnwidth}{!}{
    \begin{tabular}{lllll}
\toprule
Model & n = 10 & n = 20 & n = 50 & n = 100 \\
\midrule
GNCA & 0.49 $\pm$ 0.00 & 0.50 $\pm$ 0.00 & 0.50 $\pm$ 0.00 & 0.50 $\pm$ 0.00 \\
Recurrent GNN & 0.97 $\pm$ 0.04 & 0.95 $\pm$ 0.06 & 0.90 $\pm$ 0.13 & 0.84 $\pm$ 0.14 \\
\midrule
\model & 0.74 $\pm$ 0.14 & 0.67 $\pm$ 0.17 & 0.63 $\pm$ 0.20 & 0.61 $\pm$ 0.20 \\
\bottomrule
\end{tabular}
}
    \label{tab:prefixsum-result}
\end{table}

\end{document}